\documentclass[conference]{IEEEtran}
\IEEEoverridecommandlockouts
\usepackage{cite}
\usepackage{amsmath,amssymb,amsfonts}
\usepackage{algorithmic}
\usepackage{url}
\usepackage{graphicx}
\usepackage{textcomp}
\usepackage[table,xcdraw]{xcolor}
\usepackage{booktabs}
\usepackage{array}
\usepackage{tabularx}
\usepackage{algorithm}
\usepackage{multirow}
\usepackage{diagbox}
\usepackage{soul}
\usepackage{colortbl}
\usepackage[T1]{fontenc}
\usepackage{times}
\usepackage{xcolor}
\usepackage[
  colorlinks=true,
  linkcolor=blue,
  urlcolor=black,
  citecolor=blue,
]{hyperref}
\def\BibTeX{{\rm B\kern-.05em{\sc i\kern-.025em b}\kern-.08em
    T\kern-.1667em\lower.7ex\hbox{E}\kern-.125emX}}

\title{Strategy-Augmented Planning for Large Language Models via Opponent Exploitation}

\author{
    Shuai Xu\textsuperscript{1,2,3,*}, 
    Sijia Cui\textsuperscript{4,5,*}\thanks{\textsuperscript{*}Equal contribution}, 
    Yanna Wang\textsuperscript{4}, 
    Bo Xu\textsuperscript{2,4,\dag}\thanks{\textsuperscript{\dag}Corresponding author}, 
    Qi Wang\textsuperscript{1} \\
    \textsuperscript{1}Nanjing University of Information Science \& Technology, Nanjing, China \\
    \textsuperscript{2}Nanjing Artificial Intelligence Research of IA, Nanjing, China \\
    \textsuperscript{3}University of Chinese Academy of Sciences, Nanjing, China \\
    \textsuperscript{4}Institute of Automation, Chinese Academy of Sciences, Beijing, China \\
    \textsuperscript{5}School of Artificial Intelligence, University of Chinese Academy of Sciences, Beijing, China \\
    \texttt{xushuai23@mails.ucas.ac.cn}
}

\begin{document}

\maketitle

\begin{abstract}
Efficiently modeling and exploiting opponents is a long-standing challenge in adversarial domains. 
Large Language Models (LLMs) trained on extensive textual data have recently demonstrated outstanding performance in general tasks, introducing new research directions for opponent modeling. 
Some studies primarily focus on directly using LLMs to generate decisions based on the elaborate prompt context that incorporates opponent descriptions, while these approaches are limited to scenarios where LLMs possess adequate domain expertise. 
To address that, we introduce a two-stage Strategy-Augmented Planning (SAP) framework that significantly enhances the opponent exploitation capabilities of LLM-based agents by utilizing a critical component,the Strategy Evaluation Network (SEN).
Specifically, in the offline stage, we construct an explicit strategy space and subsequently collect strategy-outcome pair data for training the SEN network. 
During the online phase, SAP dynamically recognizes the opponent's strategies and greedily exploits them by searching best response strategy on the well-trained SEN, finally translating strategy to a course of actions by carefully designed prompts. 
Experimental results show that SAP exhibits robust generalization capabilities, allowing it to perform effectively not only against previously encountered opponent strategies but also against novel, unseen strategies. In the MicroRTS environment, SAP achieves a 85.35\% performance improvement over baseline methods and matches the competitiveness of reinforcement learning approaches against state-of-the-art (SOTA) rule-based AI.
Our code is available at \texttt{\url{https://github.com/hsushuai/SAP}}.
\end{abstract}


\section{Introduction}

Designing an artificial intelligence system has always been a challenge in academia, especially in complex game scenarios.
Games~\cite{vinyals2017starcraft, huang2021gym, ontanon2013microrts} have long been a platform for testing and evaluating agent systems, drawing sustained attention from the academic community. 
A key challenge lies in the fact that the opponent's decisions influence the state of the environment, which in turn affects the payoff of our decisions.
Therefore, it is crucial for an intelligent agent to effectively model the opponent, particularly by predicting the adversary’s actions and further inferring their underlying intentions~\cite{he2016opponent}.
Previous opponent modeling methods either require extensive domain knowledge~\cite{10.5555/3020336.3020403, ganzfried2011game} or consume substantial computational resources, including data collection and neural network fitting~\cite{he2016opponent, nashed2022survey, huang2024robust}.
Motivated by these approaches and existing limitations, we aim to develop an effective opponent-modeling and opponent-exploitation framework.

In the field of natural language processing (NLP), large language models (LLMs) trained on extensive textual data have rapidly advanced, demonstrating exceptional capabilities in general tasks~\cite{brown2020language, kojima2022large, ma2023large}.
Some researchers~\cite{zhang2024agent_pro} explicitly model the opponent strategy (e.g., “player-1 is relatively conservative”), and then progressively optimize the behavioral policy by feedback-based reflection, enabling LLM-based agent to outperform traditional models in board games.
Similarly, some studies~\cite{xu2023exploring} explore applying LLMs to multi-agent communication game scenarios, where agent requires deceiving opponents, identifying their hidden identities and true intentions, and ultimately achieving victory for the team. 
For more complex real-time strategy (RTS) games, the CoS~\cite{ma2023large} method analyzes the opponent's strategy as in-context and directly generates macro plans, translated to micro action queue by rule-based scripts. 
The superior performance can be partly attributed to the LLM's high familiarity with the scenarios of StarCraft II~\cite{ma2023large}. As a result, this approach is limited to scenarios where the LLM possesses sufficient domain knowledge.

To address these challenges, we build explicit environment-specific strategy space and propose SAP framework, a two-stage opponent exploitation approach. 
In the offline stage, we aim to fit a strategy evaluation network (SEN) used to predict the win probability between two competitive strategies. Specifically, we first utilize LLM to generate a diverse set of strategies within the strategy space. 
Next, we gather the competition outcomes by having each pair of strategies compete within a defined environment. Finally, we feed the training data, including paired strategies and their outcome, into the SEN network for the fitting process.
In the online exploitation phase, we employ an LLM-based recognizer to identify the opponent's strategy. Following this recognition, we perform a greedy search for the optimal response strategy based on the outcomes predicted by the SEN model. 
To handle the uncertainty and variability of the opponent's strategy, SAP adaptively re-identifies and searches for the optimal counter-strategy in every $k$ environment steps. 

\begin{figure*}
    \centering
    \includegraphics[width=\linewidth]{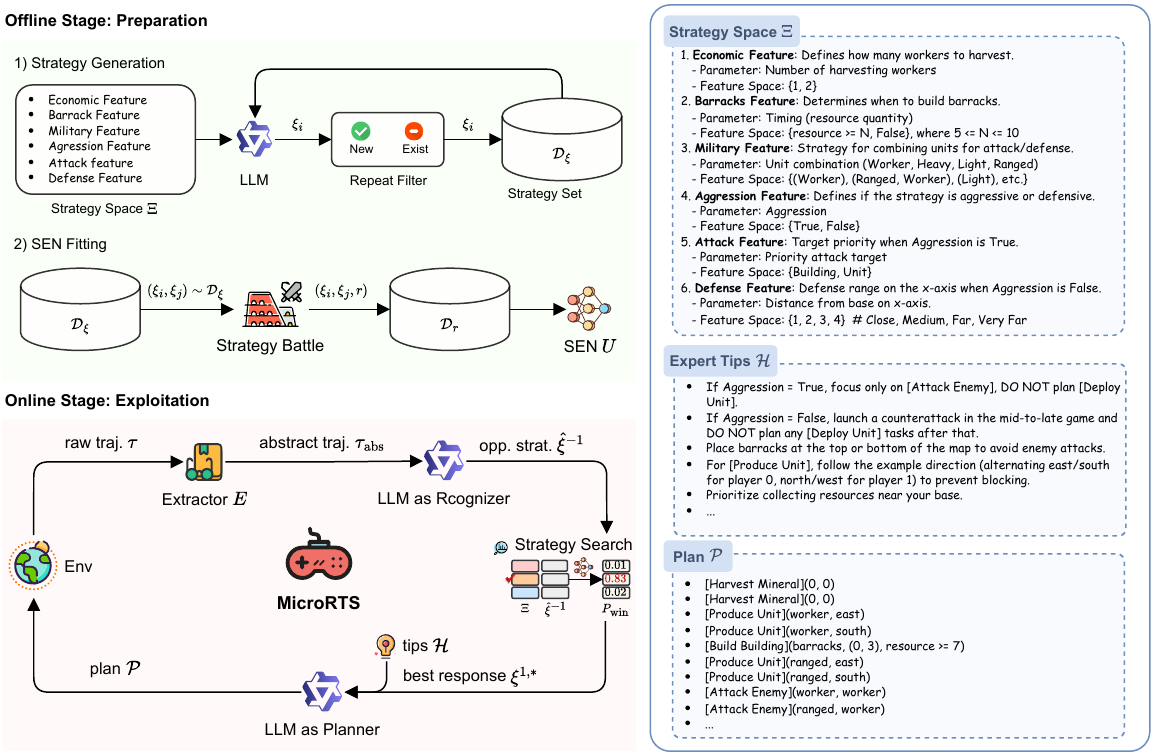}
    \caption{The Framework of SAP. In the offline stage 1, LLMs are instructed to generate diverse strategies based on the predefined strategy space $\Xi$. In offline stage 2, the generated strategy set $\mathcal{D}_{\xi}$ is extended into a strategy battle results set $\mathcal{D}_r$ by evaluating combat pairs from $\mathcal{D_{\xi}}$ in a realistic environment. This results set $\mathcal{D}_r$ is then used to train the Strategy Evaluation Network (SEN) $U$. In the online stage, the raw trajectory $\tau$ returned from the environment is processed by the trajectory extractor $E$ to produce a summarized trajectory $\tau_{\text{abs}}$, which is passed to the LLM to identify the opponent's strategy $\hat{\xi}^{-1}$. The best response strategy $\xi^{1,*}$ is then searched using SEN $U$ to guide the LLM in making informed plans, augmented by expert tips $\mathcal{H}$ to bridge the gap between strategy formulation and planning.}
    \label{fig:framework}
\end{figure*}

Our main contributions are threefold:
\begin{enumerate}
    \item \textbf{Construction of Explicit Strategy Space and SEN}: We build a well-defined strategy space and train a strategy evaluation network based on collected data.
    \item \textbf{Opponent Strategy Recognition and Exploitation}: By recognizing opponent strategies in real-time, we exploit it and greedily search best response strategy, optimizing the LLM-based agent's plan through in-context learning.
    \item \textbf{Evaluation of SAP in MicroRTS}: We conduct comprehensive experiments in the MicroRTS to validate the effectiveness of the proposed SAP method. Furthermore, the flexibility and generalizability of SAP make it applicable to other strategic planning domains. 
\end{enumerate}

\section{Related Work}

\subsection{LLMs for Planning}

LLMs have gained significant attention as a promising approach to Artificial General Intelligence (AGI) and agent systems \cite{wang2024agent_survey}. Recent work on LLM-based planning \cite{huang2024planning_survey} primarily focuses on task decomposition, search, and the integration of external planners, among other aspects. For task decomposition, methods such as HuggingGPT \cite{shen2024hugginggpt} and Plan-and-Solve \cite{wang2023plan_solve} decompose tasks in a single step, while Chain-of-Thought (CoT) \cite{wei2022cot} and ReAct \cite{yao2022react} break down tasks iteratively. Despite their impressive reasoning abilities, LLMs often generate suboptimal plans. To address this limitation, Tree-of-Thought \cite{yao2024tot} employs BFS/DFS for optimal path search, while LLM-MCTS \cite{zhao2024llm_mcts} and RAP \cite{hao2023rap} combine LLM reasoning with Monte Carlo Tree Search (MCTS), and LLM A* \cite{xiao2023llm_astar} adopts a heuristic cost function to introduce the A* algorithm into the LLM planning process. Additionally, the PDDL model \cite{aeronautiques1998pddl} uses symbolic reasoning to identify the optimal path from an initial state to a target. Approaches such as LLM+P \cite{liu2023llm+p}, LLM-DP \cite{dagan2023llm_dp}, and LLM-PDDL \cite{guan2023llm+pddl} explore combining LLMs with PDDL for enhanced planning. Unlike existing approaches, our framework leverages the best-response strategy to optimize the prompting of LLMs, facilitating the development of more reasonable and effective planning processes.
\subsection{Opponent Modeling}

Opponent modeling is a well-established research topic, with numerous studies addressing it from game theory and model-based perspectives. Recent approaches include learning the opponent's strategy space via embedding representations \cite{jing2023emb3}, inferring the opponent's strategy online using Bayesian methods \cite{lv2023online3}, and predicting and adapting to the opponent's behavior by modeling their beliefs \cite{raileanu2018tom_belif}, goals \cite{rabinowitz2018tom_goal}, and actions \cite{von2017tom_action} using theory of mind (ToM). Some works have also explored applying LLMs in opponent modeling. For instance, Agent-Pro \cite{zhang2024agent_pro} models the opponent's play style (e.g., aggressive or conservative) by establishing and maintaining the agent's world beliefs, while BIDDER \cite{zhang2024bidder} introduces bi-directional deliberate reasoning, which combines past trajectories and potential future actions of the opponent to inform decision-making. In comparison, our approach explicitly maps the strategy to a defined strategy space, improving the interpretability of the planning process in LLMs. Moreover, the low-dimensional nature of the strategy features dramatically cuts down on both the training cost and network complexity.

\section{Method}

To enhance the planning and opponent exploitation capabilities of LLM-based agents, we propose a novel framework, termed strategy-augmented planning (SAP). SAP enables agents to maximize exploitation in competitive environments by explicitly modeling strategies and constructing a strategy evaluation network (SEN). It facilitates planning by identifying the opponent's strategy and deploying a corresponding response strategy that maximizes the agent's win rate.

As shown in Fig.~\ref{fig:framework}, SAP operates in two main stages: (1) Offline preparation. LLMs generate a set of strategies based on a predefined strategy space. These strategies are evaluated through realistic battles within the environment, producing result data from competitive interactions. This data is then used to train the SEN, which predicts the probabilities of winning for both the agent's and the opponent's strategies. (2) Online exploitation. Based on the observed trajectory of the opponent, SAP identifies their strategy and searches for the optimal response strategy that maximizes the agent's probability of winning. The optimized strategy guides the LLM in formulating detailed plans. Additionally, to address potential gaps between high-level strategies and granular planning, carefully designed expert tips are employed to align strategy formulation with actionable planning steps.

\subsection{Preliminaries}

We model the planning task in a RTS game as a Markov Decision Process (MDP), defined by the tuple $(\mathcal{S}, \mathcal{A}, P, R, \gamma, \mathcal{O})$, where $\mathcal{S}$ is the state space, $\mathcal{A}$ is the action space, $P: \mathcal{S} \times \mathcal{A} \times \mathcal{S} \rightarrow [0,1]$ is the state transition probability, $R: \mathcal{S} \times \mathcal{A} \rightarrow \mathbb{R}$ is the reward function, and $\gamma$ is the discount factor.

To reduce the high computational cost of using LLMs in long-horizon games, we utilize abstract actions in the environment. These abstract actions encapsulate atomic actions $a \in \mathcal{A}$ into parameterized forms $(A, \theta)$, where $A \mapsto \{a_0, a_1, \dots\}$ is the abstract action and $\theta \in \Theta$ represents the parameters. This approach reduces the frequency of LLM calls. Our goal is to use the LLM $\Phi_{\text{LLM}}$ to plan an optimal sequence of abstract actions $\mathcal{P} = \{A_1(\theta_1), A_2(\theta_2), \dots\}$ that maximizes the expected payoff in a competitive interaction with an opponent:

\begin{equation}
    \mathcal{P}^{*} = \arg\max_{\mathcal{P}} \mathbb{E} \left[ \sum_t \gamma^t R\left(\mathcal{P}_t, s_t \right)\right]
\end{equation}

where $\mathcal{P}_t$ denotes the plan of the agent at time step $t$, and $s_t$ represents the state of the environment at time step $t$.

\begin{table}[htbp]
\centering
\caption{List of definitions in the paper.}
\label{tab:definitions}
\resizebox{\linewidth}{!}{
\begin{tabular}{p{0.25\linewidth} p{0.75\linewidth}}
\toprule
    \textbf{Name} & \textbf{Definition} \\
    \midrule
    Strategy $\xi$ & Explicit, verbalizable strategy, one strategy corresponding to several plan possibilities \\
    Plan $\mathcal{P}$ & An executable plan consisting of a series of abstract actions \\
    Abstract Action $A$ & Contains multiple atomic actions \\
    Atomic Action $a$ & The lowest level of action \\
\bottomrule
\end{tabular}
}
\end{table}

\subsection{Offline: Preparation}

In the offline phase, our primary goal is to construct the strategy evaluation network (SEN), denoted as $U$, which predicts the outcomes of competitive interactions between two strategies:

\begin{equation}
    U(\xi^1, \xi^{-1}) = r
\end{equation}

where $\xi^1$ is the agent’s strategy, $\xi^{-1}$ is the opponent’s strategy, and $r$ is the final outcome, i.e., the probability that $\xi^1$ wins.

\subsubsection{Strategy Generation}

To enable LLMs to effectively adapt to unfamiliar environments, we define strategies at the semantic level, explicitly representing them along orthogonal dimensions, as shown in Fig. \ref{fig:framework}. These semantic strategies guide LLMs in generating coherent plans without the need for additional fine-tuning. Specifically, we map each strategy to an explicit strategy space $\Xi$ and prompt LLMs $\Phi_{\text{LLM}}$ to iteratively generate strategies, creating a strategy library, $\mathcal{D_\xi}$, where each new strategy is generated with reference to previously generated ones to avoid duplication as much as possible.

\begin{equation}
\label{eq:strat_gen}
    \mathcal{D}_{\xi} = \{\xi_i | \xi_i = \Phi_{\text{LLM}}(\mathcal{I},\Xi, s_0, \{\xi_j\}_{j < i})\}
\end{equation}

where $\mathcal{I}$ represents the basic environment information, which includes the basic settings and rules of the game, $s_0\in \mathcal{S}$ is the initial state, and $\{\xi_j\}_{j < i}$ denotes the set of previously generated strategies. Additionally, we apply duplicate filtering to the strategies generated by LLMs to ensure the uniqueness of the strategies in the strategy library.

\subsubsection{SEN Fitting}
Although the strategy features we designed are language-based, they can be mapped to discrete numerical vector spaces through rule-based relationships. This mapping facilitates the fitting of the strategy evaluation network because it significantly reduces the feature dimension compared to that of the deep neural network (DNN) used for policy. To train the SEN $U$, strategies from the strategy library $\mathcal{D}_{\xi}$ are paired and evaluated through real competitive interactions, generating result data $\mathcal{D}_r = \{ (\xi_i, \xi_j, r_{ij}) \}_{i,j}$. During these interactions, LLMs generate plans consisting of abstract actions based on the respective strategies. Furthermore, to bridge the gap between strategy and plan, we design expert tips to address this discrepancy. Note that these expert tips $\mathcal{H}$ are also utilized during the online planning phase, as detailed in Section \ref{sec:plan_based_strat}.

\begin{align}
    \label{eq:battle}
    r_{ij} &= \mathbb{E}_{(\xi_i, \xi_j) \sim \mathcal{D}_{\xi}} \left[ \sum^N R \left( \mathcal{P}^1 \left( \xi_i \right), \mathcal{P}^{-1} \left( \xi_j \right) \right) \right] \\
    \label{eq:plan}
    \mathcal{P}(\xi) &= \Phi_{\text{LLM}} \left( o_t \mid \xi, \mathcal{I}, \mathcal{H} \right)
\end{align}

where $r_{ij}$ represents the expected result of the combat between strategy $\xi_i$ and $\xi_j$ in $N$ episodes, and $\mathcal{P}^1$ and $\mathcal{P}^{-1}$ are the plans of the two players, generated by the LLM $\Phi_{\text{LLM}}$. Due to the symmetry of the game, we disregard the impact of player positions (i.e., $r_{ij}=r_{ji}$), as this effect can be eliminated by rotating the map. The SEN is optimized using binary cross-entropy (BCE) loss.

\begin{align}
\label{eq:sen_loss}
    \mathcal{L}_r &= -\mathbb{E}_{(\xi_i^1, \xi_j^{-1}, r_{ij}) \sim \mathcal{D}_r} \bigg[ r_{ij} \log \Big( U\big( r_{ij} \mid \xi_i^1, \xi_j^{-1} \big) \Big) \\
    &\quad + (1 - r_{ij}) \log \Big( 1 - U\big( r_{ij} \mid \xi_i^1, \xi_j^{-1} \big) \Big) \bigg] \notag
\end{align}

Algorithm \ref{alg:offline_alg} presents the overall process of the offline stage, where \eqref{eq:strat_gen} is used to generate strategies, \eqref{eq:battle} is used to obtain the results between two strategies, and the loss in \eqref{eq:sen_loss} is used to train the SEN.

\begin{algorithm}
    \caption{Offline Preparation}
    \label{alg:offline_alg}
    \begin{algorithmic}[1]
        \STATE \textcolor{gray}{/* Offline Stage 1: Strategy Generation */}
        \STATE Initialize strategy set $\mathcal{D}_{\xi} \gets \emptyset$, and set its size to $K$.
        \FOR{$i = 1 \text{ to } K$}
        \STATE $\xi_i \gets \text{none}$
        \WHILE{$\xi_i = \text{none}$ or $\xi_i \in \mathcal{D}_{\xi}$}
        \STATE Generate an explicit strategy $\xi_i \in \Xi$ using \eqref{eq:strat_gen}
        \ENDWHILE
        \STATE Update $\mathcal{D}_{\xi} \gets \mathcal{D}_{\xi} \cup \{\xi_i\}$
        \ENDFOR
        \STATE \textcolor{gray}{/* Offline Stage 2: SEN Fitting */}
        \STATE Initialize SEN $U_{\theta_r}$ with parameters $\theta_r$;
        \STATE Set batch size $B$;
        \STATE Initialize battle result dataset $\mathcal{D}_r \gets \emptyset$.
        \FOR{$(\xi_i, \xi_j) \in \mathcal{D}_{\xi}$}
        \STATE Obtain the result $r_{ij}$ using \eqref{eq:battle}
        \STATE Update $\mathcal{D}_r \gets \mathcal{D}_r \cup \{(\xi_i, \xi_j, r_{ij})\}$.
        \ENDFOR
        \WHILE{not converged}
        \STATE Sample a batch $\mathcal{B} = \{(\xi_i, \xi_j, r_{ij})\}_l \sim \mathcal{D}_r$
        \STATE Update $\theta_r$ using \eqref{eq:sen_loss}.
        \ENDWHILE
    \end{algorithmic}
\end{algorithm}

\subsection{Online: Exploitation}

In the online phase, SAP aims to recognize the opponent's strategy and improve the probability of winning the competition by maximizing the exploitation of the opponent's strategy. Specifically, we construct an LLM-based strategy recognizer, $V : \tau \mapsto \Xi$, for greedy opponent exploitation.

\begin{equation}
\label{eq:search_strat}
    \xi^{1,*} = \arg\max_{\xi^1 \sim \Xi} \mathbb{E} \left[U\left(\xi^{1}, \xi^{-1}\right) \right]
\end{equation}

where $\tau = \{(o^1_0, a^1_0, a^{-1}_0), (o^1_1, a^1_1, a^{-1}_1), \dots \}$ is the trajectory of both the player and the opponent, based on the player's observation.

\subsubsection{Opponent Recognition}

In RTS environments, agents need to control a large number of units, and the game time steps are long. As a result, the original trajectories may exceed the context length limit of the LLM. To address this issue, we designed a trajectory summarizer based on heuristic rules, $E: \tau \mapsto \tau_{\text{abs}}$.

\begin{equation}
    V = E \circ \Phi_{\text{LLM}} : \tau \mapsto \hat \xi^{-1}
\end{equation}

It extracts summaries of the corresponding aspects according to the strategy dimension features. For example, for economic features, we count the number of \texttt{harvest} actions performed by the player's workers based on the trajectory; for attack or defense features, we track the positions of each player unit when issuing an \texttt{attack} action and the attack target, or the frequency with which all units appear in different positions. The key information from the extracted trajectory then prompts LLMs $\Phi_{\text{LLM}}$ to analyze and recognize the opponent's strategy.

\begin{equation}
\label{eq:strat_recog}
    \hat \xi^{-1} = \Phi_{\text{LLM}}\left(\Xi, E\left(\tau\right)\right)
\end{equation}

\subsubsection{Exploitation and Grounding}
\label{sec:plan_based_strat}

After identifying the opponent's strategy, we employ a greedy approach to maximize exploitation. To enhance the generalization ability of SAP, as shown in \eqref{eq:search_strat}, we expand our exploration to the entire strategy space $\Xi$, rather than restricting it to the offline-generated strategy library $\mathcal{D}_{\xi}$. This approach involves traversing all strategies to identify the optimal response strategy. Our goal is to identify a counter-strategy in strategy space $\Xi$ that maximizes the output of the SEN $U$, given the opponent's strategy $\hat{\xi}^{-1}$. Moreover, due to the gap between macro-strategy and micro-planning, we introduce expert tips $\mathcal{H}$ to address any discrepancies in the interpretation of semantic strategies when LLMs generate plans. These expert tips provide detailed descriptions of each strategy feature and the corresponding planning abstract action space, enabling LLMs to quickly adapt to unfamiliar environments. For example, ``If the Aggression Feature is set to True, plan more [Attack Enemy] abstract actions.''

The online procedure is listed in Algorithm \ref{alg:online_alg}, where \eqref{eq:strat_recog} is used to recognize the opponent strategy, while \eqref{eq:search_strat} is used to search for the optimal response strategy.

\begin{algorithm}
    \caption{Online Exploitation}
    \label{alg:online_alg}
    \begin{algorithmic}[1]
        \STATE Strategy Evaluation Network $U$, initial state $s_0$, and state transition probability $P$ are given. Set the current time step $t = 0$ and the plan update interval $k$.
        \WHILE{$s_t$ is not a terminal state}
        \IF{$t \mod k = 0$}
        \STATE $\tau \gets \{(o_0,a_0^1,a_0^{-1}), \dots, (o_{t-1},a_{t-1}^{1},a_{t-1}^{-1})\}$
        \STATE Recognize opponent strategy $\hat{\xi}^{-1}$ using \eqref{eq:strat_recog}.
        \STATE Search for the best response strategy $\xi^{1,*}$ using \eqref{eq:search_strat}.
        \STATE Generate a plan $\mathcal{P}$ using \eqref{eq:plan}.
        \ENDIF
        \STATE $t \gets t + 1$
        \STATE $s_{t} \gets P(s_t | s_{t-1}, \mathcal{P})$
        \ENDWHILE
    \end{algorithmic}
\end{algorithm}

\section{Experiments}
\subsection{Experimental Setup}
\subsubsection{Environment}

\begin{table}[htbp]
    \centering
    \caption{The abstract action space in microRTS.}
    \label{tab:skill_space}
    \scriptsize
    \resizebox{\linewidth}{!}{
    \begin{tabular}{@{}p{0.2\linewidth}p{0.3\linewidth}p{0.5\linewidth}@{}}
        \toprule
         \textbf{Abstract Action} & \textbf{Atomic Actions Included} & \textbf{Description} \\
         \midrule
         Deploy Unit     &  move, noop            & Moves a unit of a specified type to a specified location. \\
         Harvest Mineral &  move, harvest, return & Move to a specified location to harvest resources and return to base. \\
         Build Building  &  move, produce         & Constructs a building of the specified type at the designated location. \\
         Produce Unit    &  produce               & Produces a specified type of unit from a specified direction. \\
         Attack Enemy    &  move, attack          & Commands a specified type of unit to attack a specified enemy type. \\
         \bottomrule
    \end{tabular}
    }
\end{table}

\begin{figure}[htbp]
    \centering
    \includegraphics[width=0.8\linewidth]{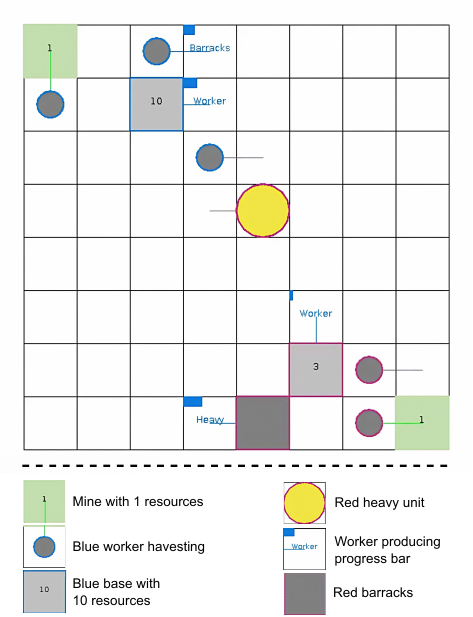}
    \caption{Example of map of a MicroRTS environment. Each unit can only move to a free horizontally or vertically adjacent cell, and each action (including moving) takes several time steps to complete.}
    \label{fig:8x8map}
\end{figure}

For evaluation, we utilize the MicroRTS environment \cite{ontanon2013microrts}, which implements abstract actions to enable the integration of LLMs, as shown in Table \ref{tab:skill_space}. Specifically, the LLM generates a plan consisting of a sequence of abstract actions for the next $N$ steps, and the environment executes the actions in the planned order. The environment is a grid-world, as shown in Fig. \ref{fig:8x8map}, where each player controls units that must gather resources, build structures, and produce more units to defeat the opponent in combat. An episode begins with each player's units starting in mirrored sections of the map and ends when only one player's units remain or a predefined step limit is reached. This environment can be configured with different map sizes, layouts, levels of partial observability, and more. Our experiments focus on the standard \texttt{basesWorkers8x8} and \texttt{basesWorkers16x16} maps, both of which provide full observability. We refer to the focal participant as the player, while the other participant is designated as the opponent.

In the offline stage, we generate 50 strategies, each of which is applicable to both players due to the symmetry of the map, and randomly divide them into 30 seen and 20 unseen strategies. Combat between the 30 seen strategies generates 900 result data points for SEN fitting. We use a 3-layer Multi-Layer Perception (MLP) as the SEN architecture, with the input concatenating of two strategy vectors, activated by ReLU, and the output as logits, passed through a sigmoid function to yield values in the range $[0, 1]$. For a detailed evaluation, refer to Section \ref{sec:sen_eval}. Furthermore, the LLMs used in the experiments are \texttt{Qwen2.5-72B-Instruct}.

\subsubsection{Baselines}

We evaluate SAP using different levels of planning methods:

\begin{itemize}
    \item Strategy-level: LLMs are prompted to generate a plan based on a fixed strategy, which includes 30 \textbf{Seen} strategies and 20 \textbf{Unseen} strategies. In each episode, the LLM generates plans based on one of these strategies and the current observations, with the final evaluation being the average performance of all the seen strategies or unseen strategies.
    \item Plan-level: LLMs generate a plan based on observations without strategy assistance, using three approaches. \textbf{Vanilla} planning is performed directly; \textbf{CoT}\cite{wei2022cot} involves appending the phrase ``Let’s think step by step:" to the last prompt; and \textbf{Tips-augmented (TA)} utilizes expert tips to optimize the LLM's output.
    \item Action-level: The agent directly outputs atomic actions for each unit at every step. The MicroRTS environment provides 13 \textbf{rule-based} bots (see Table \ref{tab:ai_bots}), including CoacAI, the winner of CoG 2020\footnote{\url{https://sites.google.com/site/micrortsaicompetition/competition-results/2020-cog-results}}. \textbf{Gridnet}\cite{huang2021gym} is a reinforcement learning (RL) method that issues actions to each cell on the map in a single step using a convolution neural network (CNN). The environment executes valid actions, ignoring those in cells without player-owned units. \textbf{Transformer}\cite{zwingenberger2023transformers} introduces a transformer-based policy for environments with variable actions.
\end{itemize}

\subsection{Results and Analysis}
\subsubsection{Main Results}

\begin{table}[htbp]
\centering
\caption{The previous MicroRTS competition bots.}
\label{tab:ai_bots}
\begin{tabular}{lll}
\toprule
    \textbf{Name} & \textbf{Category} & \textbf{Best result} \\
    \midrule
    CoacAI & Scripted & 1st place in 2020 \\
    Tiamat & MCTS-based & 1st place in 2018 \\
    MixedBot & MCTS-based & 2nd place in 2019 \\
    Droplet & MCTS-based & 3rd place in 2019 \\
    Izanagi & MCTS-based & 4th place in 2019 \\
    Rojo & MCTS-based & 5th place in 2020 \\
    LightRush & Scripted & 6th place in 2020 \\
    GuidedRojoA3N & MCTS-based & 7th place in 2020 \\
    WorkerRush & Scripted & 8th place in 2020 \\
    NaiveMCTS & MCTS-based & 9th place in 2020 \\
    RandomBiasedAI & Scripted & 10th place in 2020 \\
    Random & Scripted & - \\
    PassiveAI & Scripted & - \\
\bottomrule
\end{tabular}
\end{table}

In Table \ref{tab:main_result}, we assess the performance of the LLM-based agent in game task planning. Our proposed approach, SAP, achieves the highest win rate across various opponent types, highlighting its robust generalization capabilities. Interestingly, SAP demonstrates a higher win rate against unseen strategies compared to seen strategies. This difference may be attributed to the seen strategy set containing a higher proportion of stronger strategies, as evidenced by their superior win rates relative to the unseen strategies. Moreover, the strategy-level approach is also significantly better than the plan-level approach, with SAP outperforming TA by 85.35\%, and seen showing a 24.00\% improvement over TA.


\begin{table}[htbp]
    \centering
    \caption{Comparison of performance across different methods and strategies. Values represent the win rate for each player against various opponents.}
    \label{tab:main_result}
    \resizebox{1\columnwidth}{!}{
    \begin{tabular}{@{}l|l|ccccccc@{}}
    \toprule
        & \textbf{\begin{tabular}[c]{@{}l@{}}\diagbox{\textbf{P}}{\textit{O}}\end{tabular}} & \textit{Vanilla} & \textit{CoT} & \textit{TA} & \textit{Seen} & \textit{Unseen} & \textit{SAP} & \textbf{Avg.} \\ \midrule
        \multirow{3}{*}{{\begin{tabular}[c]{@{}l@{}}Plan \\ Level\end{tabular}}}
        & \textbf{Vanilla}   & -  & \cellcolor[HTML]{E67C73}0.00\%  & \cellcolor[HTML]{E67C73}0.00\%  & \cellcolor[HTML]{E67C73}0.00\%  & \cellcolor[HTML]{E67C73}0.00\%  & \cellcolor[HTML]{E67C73}0.00\%  & \cellcolor[HTML]{E67C73}0.00\%  \\
        & \textbf{CoT}       & \cellcolor[HTML]{57BB8A}100.00\%  & -  & \cellcolor[HTML]{E67C73}0.00\%  & \cellcolor[HTML]{E67C73}6.67\%  & \cellcolor[HTML]{E98D85}20.00\%  & \cellcolor[HTML]{E67C73}0.00\%  & \cellcolor[HTML]{F8DEDC}25.33\%  \\
        & \textbf{TA}      & \cellcolor[HTML]{57BB8A}100.00\%  & \cellcolor[HTML]{57BB8A}100.00\%  & -  & \cellcolor[HTML]{E98D85}16.67\%  & \cellcolor[HTML]{FCF1F1}45.00\%  & \cellcolor[HTML]{E67C73}0.00\%  & \cellcolor[HTML]{F4FBF7}52.33\%  \\
        \midrule
        \multirow{3}{*}{{\begin{tabular}[c]{@{}l@{}}Strategy \\ Level\end{tabular}}} 
        & \textbf{Seen}      & \cellcolor[HTML]{57BB8A}96.67\%  & \cellcolor[HTML]{57BB8A}93.33\%  & \cellcolor[HTML]{BCE4D1}70.00\%  & -  & \cellcolor[HTML]{F4FBF7}53.33\%  & \cellcolor[HTML]{E67C73}10.00\%  & \cellcolor[HTML]{BCE4D1}\underline{64.67\%}  \\
        & \textbf{Unseen}    & \cellcolor[HTML]{57BB8A}95.00\%  & \cellcolor[HTML]{6EC59A}80.00\%  & \cellcolor[HTML]{FFFFFF}50.00\%  & \cellcolor[HTML]{FCF1F1}37.50\%  & -  & \cellcolor[HTML]{E67C73}5.00\%  & \cellcolor[HTML]{F4FBF7}53.50\%  \\
        & \textbf{SAP}       & \cellcolor[HTML]{57BB8A}100.00\%  & \cellcolor[HTML]{57BB8A}100.00\%  & \cellcolor[HTML]{57BB8A}100.00\%  & \cellcolor[HTML]{57BB8A}90.00\%  & \cellcolor[HTML]{57BB8A}95.00\%  & -  & \cellcolor[HTML]{57BB8A}\textbf{97.00\%}  \\
    \bottomrule
    \end{tabular}
    }
\end{table}

\begin{figure}[htbp]
    \centering
    \includegraphics[width=\linewidth]{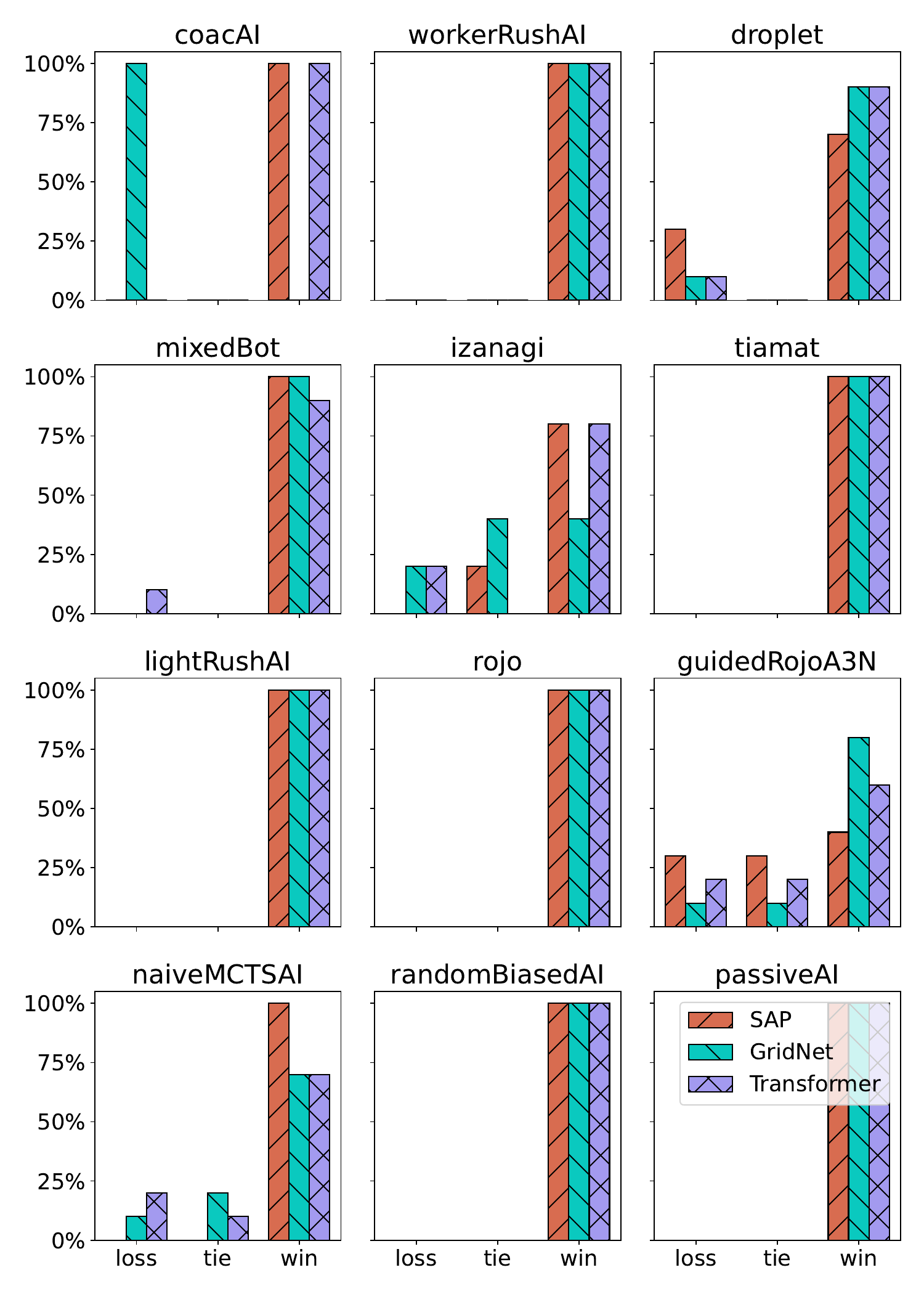}
    \caption{Results against scripted AIs. The y-axis represents the rates of wins, ties, and losses against the AIs listed in Table \ref{tab:ai_bots}. Results for the Random bot are excluded for presentation.}
    \label{fig:against_bots}
\end{figure}

As shown in Fig. \ref{fig:against_bots}, we adopt the setup from \cite{huang2021gym, zwingenberger2023transformers} and compare SAP with RL-based methods by evaluating the performance of 13 scripted AIs in the MicroRTS environment. Notably, these scripted AIs and RL-based methods plan at the action level, i.e., they plan atomic actions at every step, whereas SAP plans abstract actions every 200 steps. Despite this difference, our method demonstrates remarkable performance. Specifically, SAP outperforms RL-based methods in competitions against mixedBot, coacAI, izanagi, and naiveMCTS. It performs slightly worse against droplet and guidedRojoA3N, and shows comparable performance to RL-based methods when competing against other rule-based AIs.

\subsubsection{SEN Evaluation}
\label{sec:sen_eval}

\begin{figure}[htbp]
    \centering
    \begin{minipage}{0.48\linewidth}
    \vspace{3pt}
    \centerline{\includegraphics[width=\textwidth]{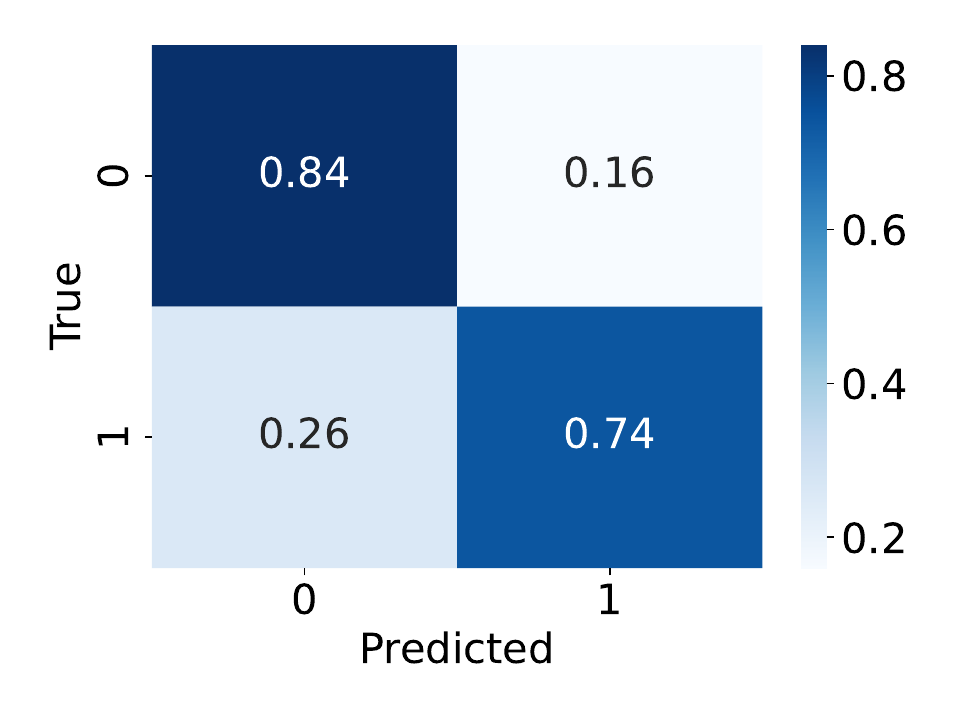}}
    \centerline{(a)}
    \end{minipage}
    \begin{minipage}{0.48\linewidth}
    \vspace{3pt}
    \centerline{\includegraphics[width=\textwidth]{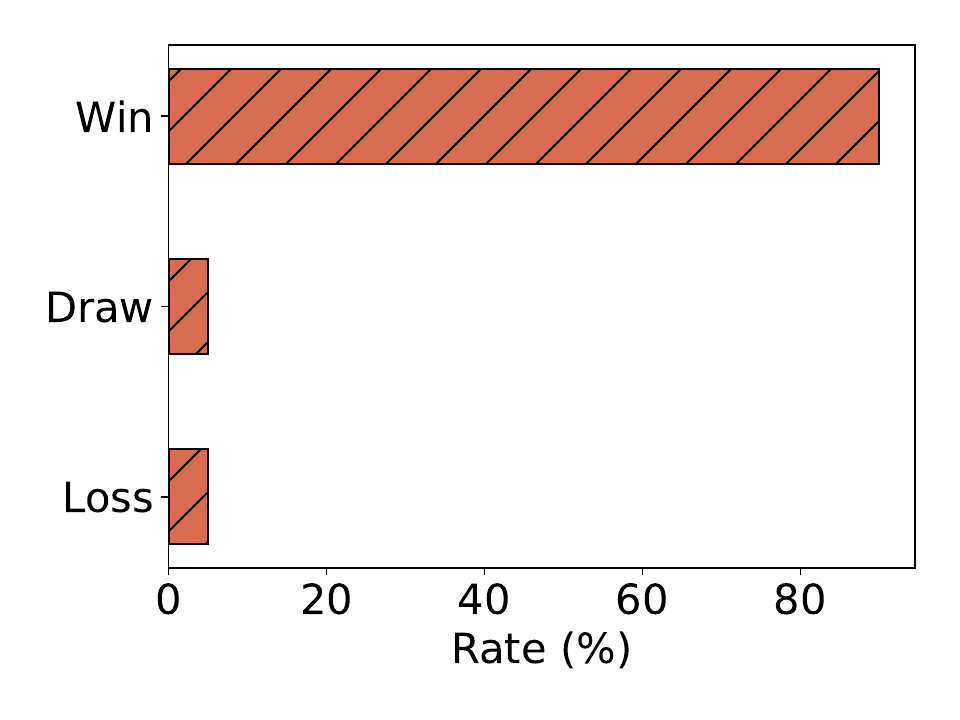}}
    \centerline{(b)}
    \end{minipage}
    \caption{SEN Evaluation. (a) Confusion matrix illustrating the accuracy of SEN predictions, where 1 represents a win and 0 represents a non-win. (b) Win, draw, and loss rates when the true opponent strategy is provided, and the strategy with the highest predicted win rate from SEN is employed.}
    \label{fig:sen_eval}
\end{figure}

SEN is a core component of SAP, and its accuracy determines the upper limit of SAP's performance. We analyze SEN's performance from two perspectives. Fig. \ref{fig:sen_eval}(a) shows the confusion matrix of SEN on the test dataset, obtained during the offline SEN fitting stage. Of particular interest is the scenario where a true value of 0 (cannot win) is predicted as 1 (can win). In other words, minimizing the False Positive (FP) rate in the upper right corner is critical. The test results indicate an FP rate of 16\%. Fig. \ref{fig:sen_eval}(b) evaluates SEN's performance in determining the best response strategy against real opponent strategies from the unseen strategy set. For each opponent strategy, the winning rate of the searched strategy is 93.3\%, with draw and loss rates of 3.3\%, further underscoring the reliability of SEN.

\subsubsection{Map Size Scaling}

\begin{table}
    \centering
    \caption{SAP win rate against different opponents on 16x16 maps.}
    \label{tab:map_scaling}
    \resizebox{1\columnwidth}{!}{
    \begin{tabular}{@{}l|cccccc@{}}
        \toprule
         \begin{tabular}[c]{@{}l@{}}\diagbox{\textbf{P}}{\textit{O}}\end{tabular} & \textit{Vanilla} & \textit{CoT} & \textit{TA} & \textit{Seen} & \textit{Unseen} & \textbf{Avg.} \\
         \midrule
         \textbf{SAP} &  100.00\% &  100.00\% &  100.00\% &  87.78\% &  88.33\% & 95.22\% \\
         \bottomrule
    \end{tabular}
    }
\end{table}

We evaluated the performance of SAP on larger maps, which involve more complex state and action spaces. As shown in Table \ref{tab:map_scaling}, SAP continues to perform well, achieving a 100.00\% win rate against Vanilla, CoT, and TA, and experiencing a 2.47\% and 7.02\% drop in win rate against seen and unseen strategy-based opponents, respectively, compared to the 8x8 maps. These results also reflect the challenges of LLM planning in more complex scenarios.

\subsubsection{Detailed Analysis}

\begin{figure}[htbp]
    \centering
    \begin{minipage}{0.8\linewidth}
    \centerline{\includegraphics[width=\textwidth]{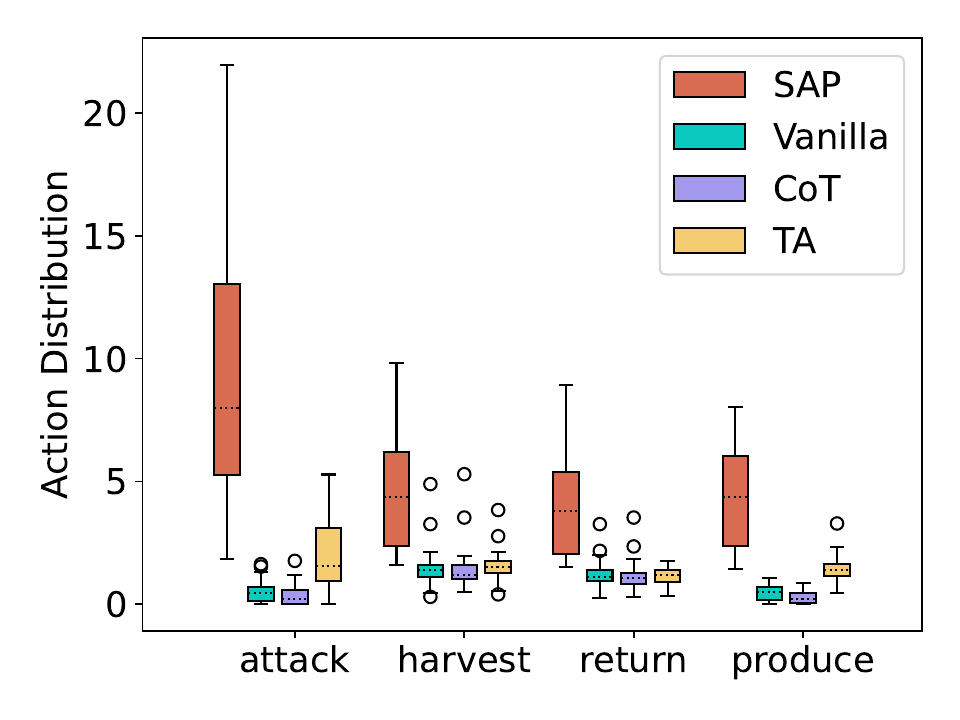}}
    \centerline{(a)}
    \vspace{3pt}
    \centerline{\includegraphics[width=\textwidth]{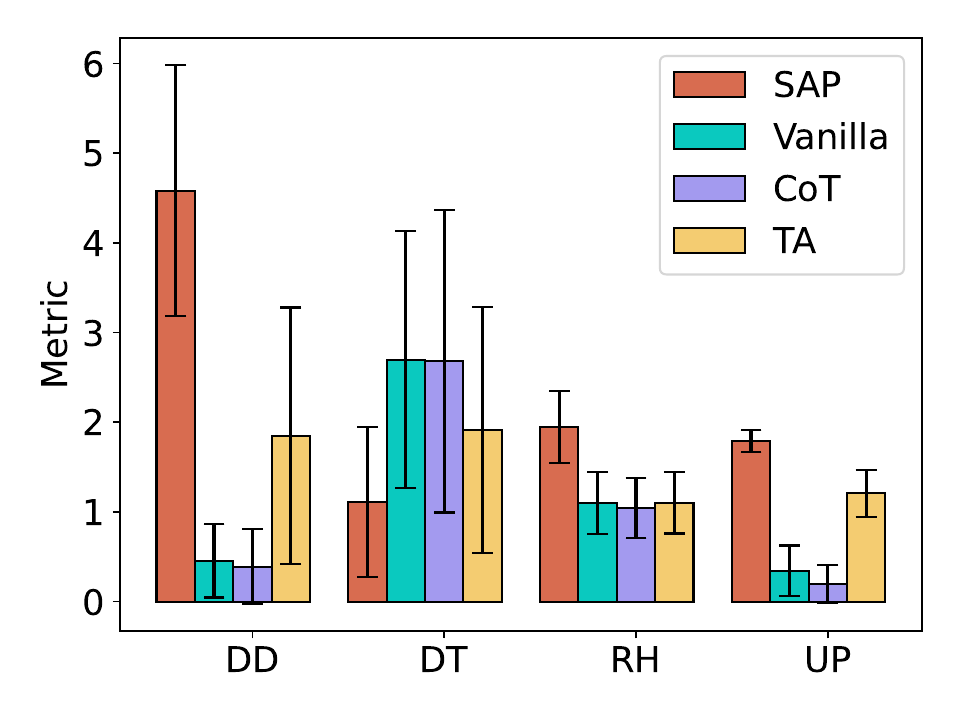}}
    \centerline{(b)}
    \end{minipage}
    \caption{Detailed Analysis. (a) Action distribution, showing the number of actions executed by all units in the game using different methods, recorded every 100 time steps; (b) Performance metrics for each player, recorded every 100 steps, excluding wins and losses.}
    \label{fig:detail_analysis}
\end{figure}

We conducted a detailed analysis of the action distribution for different methods in MicroRTS. From Fig. \ref{fig:detail_analysis}(a), it can be observed that the action frequency of SAP is significantly higher than that of other methods, primarily due to the higher frequency of the produce action. This indicates that SAP produced more units, resulting in an overall higher action frequency, particularly for attack actions, which are crucial for winning the game. Additionally, the attack frequency of TA is higher than that of Vanilla and CoT, likely due to the more aggressive expert tips utilized by TA.

Fig. \ref{fig:detail_analysis}(b) highlights other performance metrics for the different methods, excluding win/loss outcomes. SAP demonstrates superior performance, including higher damage dealt (DD), resources harvested (RH), units produced (UP), and lower damage taken (DT). These key metrics correlate with the action distribution observed in Fig. \ref{fig:detail_analysis}(a): higher damage output corresponds to more frequent attack actions, greater resource collection aligns with increased harvest and return actions, and higher unit production reflects more frequent produce actions.

\subsubsection{Ablation Studies}

\begin{table}[htbp]
    \centering
    \caption{Win rates of ablation methods against different opponent types.}
    \label{tab:ablation}
    \begin{tabular}{c|ccc}
    \toprule
        \textbf{\begin{tabular}[c]{@{}l@{}}\diagbox{\textbf{\scriptsize Player}}{\textit{\scriptsize Opponent}}\end{tabular}} & \textit{Seen} & \textit{Unseen} & \textbf{Avg.} \\
        \midrule
        \textbf{SAP}          & \textbf{90.00\%} & \textbf{95.00\%} & \textbf{92.50}\% \\
        \textbf{SAP-EPE}      & \underline{88.89\%} & \underline{91.67\%} & \underline{90.28}\% \\
        \textbf{SAP w/o SEN}  & 78.89\% & \underline{91.67\%} & 85.28\% \\
        \textbf{SAP w/o tips} & 43.33\% & 66.67\% & 55.00\% \\
    \bottomrule
    \end{tabular}
\end{table}

Table \ref{tab:ablation} shows the role of key components in SAP. SAP-EPE refers to SAP exploit per episode, where the agent updates its strategy for the next episode by utilizing the opponent's trajectory after each episode, leading to a more stable strategy. SAP w/o SEN indicates that SAP does not use SEN to search for the best response based on the opponent's strategy, but instead relies on LLM to generate the best response. SAP w/o tips means that SAP lacks expert tips to guide the transition from strategy to plans during the planning phase.

In Table \ref{tab:ablation}, employing shorter intervals between strategy updates, which promotes more greedy exploitation of the opponent, significantly enhances SAP’s adaptability to the opponent’s strategies. However, when relying solely on the LLM for opponent exploitation, a 7.81\% drop in performance is observed, underscoring the enhanced expressiveness offered by SEN. Specifically, the complete SAP achieves a 68.18\% improvement over its counterpart without expert tips, highlighting the challenges faced by the LLM in interpreting strategies and specific plans in unfamiliar environments.

\section{Conclusion}

We introduced SAP, a novel framework that grounds LLMs in complex game environments. This framework includes offline diverse strategy generation, strategy evaluation network fitting, as well as online opponent strategy recognition and optimal response strategy search. By designing an explicit strategy space, the LLM is more effectively grounded in new environments. Additionally, the reduced strategy dimension simplifies the training of the strategy evaluation network, enabling it to fit the training dataset well with minimal data. Experimental results demonstrate that our method not only significantly enhances the planning capabilities of LLMs in game confrontation scenarios, but also exhibits robust generalization.

As SAP engages in RTS games through interval planning, larger maps introduce more complex situational changes. This makes it challenging to adapt to the opponent's strategies by simply planning for a fixed future time period, as shown in Table \ref{tab:map_scaling}. Furthermore, since optimal strategies may vary across different maps, this limits the generalization capability of the SEN, requiring the training of distinct SENs for each map. Future directions for our research include enhancing the framework's capabilities for multi-player scenarios and more complex environments.

\bibliographystyle{IEEEtran}
\bibliography{references}

\begin{thebibliography}{10}
\providecommand{\url}[1]{#1}
\csname url@samestyle\endcsname
\providecommand{\newblock}{\relax}
\providecommand{\bibinfo}[2]{#2}
\providecommand{\BIBentrySTDinterwordspacing}{\spaceskip=0pt\relax}
\providecommand{\BIBentryALTinterwordstretchfactor}{4}
\providecommand{\BIBentryALTinterwordspacing}{\spaceskip=\fontdimen2\font plus
\BIBentryALTinterwordstretchfactor\fontdimen3\font minus \fontdimen4\font\relax}
\providecommand{\BIBforeignlanguage}[2]{{%
\expandafter\ifx\csname l@#1\endcsname\relax
\typeout{** WARNING: IEEEtran.bst: No hyphenation pattern has been}%
\typeout{** loaded for the language `#1'. Using the pattern for}%
\typeout{** the default language instead.}%
\else
\language=\csname l@#1\endcsname
\fi
#2}}
\providecommand{\BIBdecl}{\relax}
\BIBdecl

\bibitem{vinyals2017starcraft}
O.~Vinyals, T.~Ewalds, S.~Bartunov, P.~Georgiev, A.~S. Vezhnevets, M.~Yeo, A.~Makhzani, H.~K{\"u}ttler, J.~Agapiou, J.~Schrittwieser \emph{et~al.}, ``Starcraft ii: A new challenge for reinforcement learning,'' \emph{arXiv preprint arXiv:1708.04782}, 2017.

\bibitem{huang2021gym}
\BIBentryALTinterwordspacing
S.~Huang, S.~Onta{\~{n}}{\'{o}}n, C.~Bamford, and L.~Grela, ``Gym-{\(\mathrm{\mu}\)}rts: Toward affordable full game real-time strategy games research with deep reinforcement learning,'' in \emph{2021 {IEEE} Conference on Games (CoG), Copenhagen, Denmark, August 17-20, 2021}.\hskip 1em plus 0.5em minus 0.4em\relax {IEEE}, 2021, pp. 671--678. [Online]. Available: \url{https://doi.org/10.1109/CoG52621.2021.9619076}
\BIBentrySTDinterwordspacing

\bibitem{ontanon2013microrts}
S.~Ontan{\'o}n, ``The combinatorial multi-armed bandit problem and its application to real-time strategy games,'' in \emph{Proceedings of the AAAI Conference on Artificial Intelligence and Interactive Digital Entertainment}, vol.~9, no.~1, 2013, pp. 58--64.

\bibitem{he2016opponent}
H.~He, J.~Boyd-Graber, K.~Kwok, and H.~Daum{\'e}~III, ``Opponent modeling in deep reinforcement learning,'' in \emph{International conference on machine learning}.\hskip 1em plus 0.5em minus 0.4em\relax PMLR, 2016, pp. 1804--1813.

\bibitem{10.5555/3020336.3020403}
F.~Southey, M.~Bowling, B.~Larson, C.~Piccione, N.~Burch, D.~Billings, and C.~Rayner, ``Bayes' bluff: opponent modelling in poker,'' in \emph{Proceedings of the Twenty-First Conference on Uncertainty in Artificial Intelligence}, ser. UAI'05.\hskip 1em plus 0.5em minus 0.4em\relax Arlington, Virginia, USA: AUAI Press, 2005, p. 550–558.

\bibitem{ganzfried2011game}
S.~Ganzfried and T.~Sandholm, ``Game theory-based opponent modeling in large imperfect-information games,'' in \emph{The 10th International Conference on Autonomous Agents and Multiagent Systems-Volume 2}, 2011, pp. 533--540.

\bibitem{nashed2022survey}
S.~Nashed and S.~Zilberstein, ``A survey of opponent modeling in adversarial domains,'' \emph{Journal of Artificial Intelligence Research}, vol.~73, pp. 277--327, 2022.

\bibitem{huang2024robust}
R.~Huang, X.~Wu, H.~Yu, Z.~Fan, H.~Fu, Q.~Fu, and W.~Yang, ``A robust and opponent-aware league training method for starcraft ii,'' \emph{Advances in Neural Information Processing Systems}, vol.~36, 2024.

\bibitem{brown2020language}
T.~Brown, B.~Mann, N.~Ryder, M.~Subbiah, J.~D. Kaplan, P.~Dhariwal, A.~Neelakantan, P.~Shyam, G.~Sastry, A.~Askell \emph{et~al.}, ``Language models are few-shot learners,'' \emph{Advances in neural information processing systems}, vol.~33, pp. 1877--1901, 2020.

\bibitem{kojima2022large}
T.~Kojima, S.~S. Gu, M.~Reid, Y.~Matsuo, and Y.~Iwasawa, ``Large language models are zero-shot reasoners,'' \emph{Advances in neural information processing systems}, vol.~35, pp. 22\,199--22\,213, 2022.

\bibitem{ma2023large}
W.~Ma, Q.~Mi, Y.~Zeng, X.~Yan, Y.~Wu, R.~Lin, H.~Zhang, and J.~Wang, ``Large language models play starcraft ii: Benchmarks and a chain of summarization approach,'' \emph{arXiv preprint arXiv:2312.11865}, 2023.

\bibitem{zhang2024agent_pro}
W.~Zhang, K.~Tang, H.~Wu, M.~Wang, Y.~Shen, G.~Hou, Z.~Tan, P.~Li, Y.~Zhuang, and W.~Lu, ``Agent-pro: Learning to evolve via policy-level reflection and optimization,'' \emph{arXiv preprint arXiv:2402.17574}, 2024.

\bibitem{xu2023exploring}
Y.~Xu, S.~Wang, P.~Li, F.~Luo, X.~Wang, W.~Liu, and Y.~Liu, ``Exploring large language models for communication games: An empirical study on werewolf,'' \emph{arXiv preprint arXiv:2309.04658}, 2023.

\bibitem{wang2024agent_survey}
L.~Wang, C.~Ma, X.~Feng, Z.~Zhang, H.~Yang, J.~Zhang, Z.~Chen, J.~Tang, X.~Chen, Y.~Lin \emph{et~al.}, ``A survey on large language model based autonomous agents,'' \emph{Frontiers of Computer Science}, vol.~18, no.~6, p. 186345, 2024.

\bibitem{huang2024planning_survey}
X.~Huang, W.~Liu, X.~Chen, X.~Wang, H.~Wang, D.~Lian, Y.~Wang, R.~Tang, and E.~Chen, ``Understanding the planning of llm agents: A survey,'' \emph{arXiv preprint arXiv:2402.02716}, 2024.

\bibitem{shen2024hugginggpt}
Y.~Shen, K.~Song, X.~Tan, D.~Li, W.~Lu, and Y.~Zhuang, ``Hugginggpt: Solving ai tasks with chatgpt and its friends in hugging face,'' \emph{Advances in Neural Information Processing Systems}, vol.~36, 2024.

\bibitem{wang2023plan_solve}
L.~Wang, W.~Xu, Y.~Lan, Z.~Hu, Y.~Lan, R.~K.-W. Lee, and E.-P. Lim, ``Plan-and-solve prompting: Improving zero-shot chain-of-thought reasoning by large language models,'' \emph{arXiv preprint arXiv:2305.04091}, 2023.

\bibitem{wei2022cot}
J.~Wei, X.~Wang, D.~Schuurmans, M.~Bosma, F.~Xia, E.~Chi, Q.~V. Le, D.~Zhou \emph{et~al.}, ``Chain-of-thought prompting elicits reasoning in large language models,'' \emph{Advances in neural information processing systems}, vol.~35, pp. 24\,824--24\,837, 2022.

\bibitem{yao2022react}
S.~Yao, J.~Zhao, D.~Yu, N.~Du, I.~Shafran, K.~Narasimhan, and Y.~Cao, ``React: Synergizing reasoning and acting in language models,'' \emph{arXiv preprint arXiv:2210.03629}, 2022.

\bibitem{yao2024tot}
S.~Yao, D.~Yu, J.~Zhao, I.~Shafran, T.~Griffiths, Y.~Cao, and K.~Narasimhan, ``Tree of thoughts: Deliberate problem solving with large language models,'' \emph{Advances in Neural Information Processing Systems}, vol.~36, 2024.

\bibitem{zhao2024llm_mcts}
Z.~Zhao, W.~S. Lee, and D.~Hsu, ``Large language models as commonsense knowledge for large-scale task planning,'' \emph{Advances in Neural Information Processing Systems}, vol.~36, 2024.

\bibitem{hao2023rap}
S.~Hao, Y.~Gu, H.~Ma, J.~J. Hong, Z.~Wang, D.~Z. Wang, and Z.~Hu, ``Reasoning with language model is planning with world model,'' \emph{arXiv preprint arXiv:2305.14992}, 2023.

\bibitem{xiao2023llm_astar}
H.~Xiao and P.~Wang, ``Llm a*: Human in the loop large language models enabled a* search for robotics,'' \emph{arXiv preprint arXiv:2312.01797}, 2023.

\bibitem{aeronautiques1998pddl}
C.~Aeronautiques, A.~Howe, C.~Knoblock, I.~D. McDermott, A.~Ram, M.~Veloso, D.~Weld, D.~W. Sri, A.~Barrett, D.~Christianson \emph{et~al.}, ``Pddl| the planning domain definition language,'' \emph{Technical Report, Tech. Rep.}, 1998.

\bibitem{liu2023llm+p}
B.~Liu, Y.~Jiang, X.~Zhang, Q.~Liu, S.~Zhang, J.~Biswas, and P.~Stone, ``Llm+ p: Empowering large language models with optimal planning proficiency,'' \emph{arXiv preprint arXiv:2304.11477}, 2023.

\bibitem{dagan2023llm_dp}
G.~Dagan, F.~Keller, and A.~Lascarides, ``Dynamic planning with a llm,'' \emph{arXiv preprint arXiv:2308.06391}, 2023.

\bibitem{guan2023llm+pddl}
L.~Guan, K.~Valmeekam, S.~Sreedharan, and S.~Kambhampati, ``Leveraging pre-trained large language models to construct and utilize world models for model-based task planning,'' \emph{Advances in Neural Information Processing Systems}, vol.~36, pp. 79\,081--79\,094, 2023.

\bibitem{jing2023emb3}
Y.~Jing, K.~Li, B.~Liu, Y.~Zang, H.~Fu, Q.~FU, J.~Xing, and J.~Cheng, ``Towards offline opponent modeling with in-context learning,'' in \emph{The Twelfth International Conference on Learning Representations}, 2023.

\bibitem{lv2023online3}
Y.~Lv, Y.~Yu, Y.~Zheng, J.~Hao, Y.~Wen, and Y.~Yu, ``Limited information opponent modeling,'' in \emph{International Conference on Artificial Neural Networks}.\hskip 1em plus 0.5em minus 0.4em\relax Springer, 2023, pp. 511--522.

\bibitem{raileanu2018tom_belif}
R.~Raileanu, E.~Denton, A.~Szlam, and R.~Fergus, ``Modeling others using oneself in multi-agent reinforcement learning,'' in \emph{International conference on machine learning}.\hskip 1em plus 0.5em minus 0.4em\relax PMLR, 2018, pp. 4257--4266.

\bibitem{rabinowitz2018tom_goal}
N.~Rabinowitz, F.~Perbet, F.~Song, C.~Zhang, S.~A. Eslami, and M.~Botvinick, ``Machine theory of mind,'' in \emph{International conference on machine learning}.\hskip 1em plus 0.5em minus 0.4em\relax PMLR, 2018, pp. 4218--4227.

\bibitem{von2017tom_action}
F.~B. Von Der~Osten, M.~Kirley, and T.~Miller, ``The minds of many: Opponent modeling in a stochastic game.'' in \emph{IJCAI}, 2017, pp. 3845--3851.

\bibitem{zhang2024bidder}
Y.~Zhang, S.~Mao, W.~Wu, Y.~Xia, T.~Ge, M.~Lan, and F.~Wei, ``Enhancing language model rationality with bi-directional deliberation reasoning,'' \emph{arXiv preprint arXiv:2407.06112}, 2024.

\bibitem{zwingenberger2023transformers}
N.~Zwingenberger, ``Transformers as policies for variable action environments,'' 2023.

\end{thebibliography}

\end{document}